\documentclass[10pt,twocolumn,letterpaper]{article}

\usepackage[pagenumbers]{cvpr} 

\usepackage[dvipsnames]{xcolor}

\def\ensclip{{Ensembled CLIP score}}
\def\consensus{{Consensus score}}
\def\ourmethod{{\textbf{ECO}}}

\definecolor{cvprblue}{rgb}{0.21,0.49,0.74}
\usepackage[pagebackref,breaklinks,colorlinks,citecolor=cvprblue]{hyperref}

\usepackage{multirow}
\usepackage{multicol}
\usepackage{float}

\def\paperID{*****} 
\def\confName{CVPR}
\def\confYear{2024}
\def\workshop{NICE}

\title{Technical Report of NICE Challenge at CVPR 2024: \\
Caption Re-ranking Evaluation Using Ensembled CLIP and Consensus Scores}
\author{Kiyoon Jeong\thanks{Equal Contribution},
Woojun Lee\footnotemark[1],
Woongchan Nam,
Minjeong Ma,
Pilsung Kang\thanks{Corresponding author}\\
School of Industrial and Management Engineering, Korea University\\
{\tt\small \{kiyoon\_jeong, woojun\_lee, woongchan\_nam, minjeong\_ma, pilsung\_kang\}@korea.ac.kr}
}

\begin{document}


\twocolumn[{
\maketitle
\begin{center}
    \captionsetup{type=figure}
    \includegraphics[height=5cm]{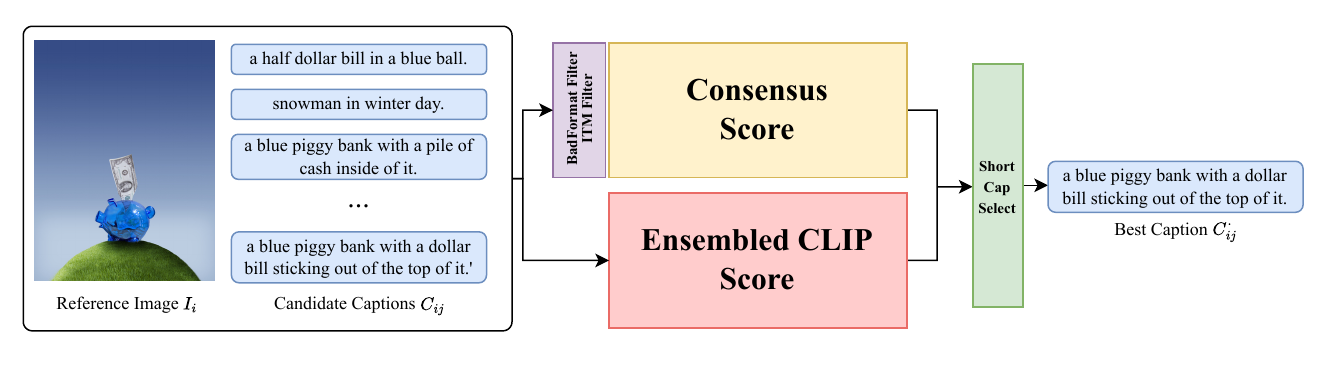}
    \captionof{figure}{Overview of \ourmethod{} (\textbf{E}nsembled \textbf{C}lip score and c\textbf{O}nsensus score) framework utilizes both the \ensclip{} and \consensus{} to select the ideal caption.}
    \label{fig:pipeline}
\end{center}
}]

\def\ensclip{{Ensembled CLIP score}}
\def\consensus{{Consensus score}}
\def\ourmethod{{\textbf{ECO}}}

\begin{abstract}
This report presents the \ourmethod{} (\textbf{E}nsembled \textbf{C}lip score and c\textbf{O}nsensus score) pipeline from team \textbf{DSBA LAB}, which is a new framework used to evaluate and rank captions for a given image. \ourmethod{} selects the most accurate caption describing image. It is made possible by combining an \ensclip{}, which considers the semantic alignment between the image and captions, with a \consensus{} that accounts for the essentialness of the captions. Using this framework, we achieved notable success in the \confName \space \confYear \space Workshop Challenge on Caption Re-ranking Evaluation at the New Frontiers for Zero-Shot Image Captioning Evaluation (\workshop). Specifically, we secured third place based on the CIDEr metric, second in both the SPICE and METEOR metrics, and first in the ROUGE-L and all BLEU Score metrics. The code and configuration for the \ourmethod{} framework are available at 
\url{https://github.com/DSBA-Lab/ECO}
.
\end{abstract} 
\def\thefootnote{*}\footnotetext{Equal Contribution.}
\def\thefootnote{$\dagger$}\footnotetext{Corresponding author.}
\section{Introduction}
\label{sec:intro}

The NICE 2024 Challenge Caption Re-ranking track is a competition that challenges participants to identify the most accurate and comprehensive caption from a set of candidate captions for a given image. The goal is to select caption that accurately and thoroughly describe an image. The NICE dataset provided for this challenge comprises 20,000 images, and about 60 captions for each image, forming a zero-shot evaluation dataset. It includes various images, candidate captions generated by different models, and undisclosed answer captions created by human annotators.

Participants in the ‘Image Caption Re-ranking’ task should choose and submit the caption they consider most appropriate for each image. Their submissions are evaluated against five undisclosed correct captions written by different human annotators, based on five metrics: CIDEr \cite{vedantam2015cider}, SPICE \cite{anderson2016spice}, METEOR\cite{banerjee2005meteor}, ROGUE-L\cite{lin2004rouge}, and BLEU\cite{papineni2002bleu}. The aim is to encourage innovative approaches to selecting captions that enhance the accuracy and depth of image descriptions.

Upon first glance, one may assume this task is similar to the image-to-text retrieval task. However, there is a distinction as the Retrieval Task involves selecting the clear answer caption among multiple candidate captions, some of which may significantly deviate from the answer. This task is evaluated using recall metrics such as recall@1, 5, and 10 \cite{salton1971smart}. On the other hand, ‘Image Caption Re-ranking’ task literally focuses on re-ranking candidate captions, which are closely aligned with the correct answers. These differences call for evaluating captions in more detail.

To develop an algorithm capable of re-ranking captions based on the quality of their descriptions of images, it was important first to establish a clear definition of what constitutes an “accurate and thorough” caption. To achieve this, we determined that an ideal caption must meet two key criteria:
\begin{quote}
    1. An ideal caption should have a high semantic alignment with the associated image.
\\
    2. An ideal caption should have a high degree of essentialness.
\end{quote}
The first criterion for a caption is that it should accurately reflect the context of the image and not include any content that is not present in the image. In other words, the caption should be semantically aligned with the image, and the more alignment there is, the better it will meet the first criterion.

The second criterion requires that the caption avoids using overly elaborate language and instead focuses on using essential expressions. This means that the caption should only include expressions that are necessary to describe the image. The more indispensable each expression is for accurately depicting the image, the better it meets the second criterion.

It is important to note that meeting only one of the criteria does not guarantee the caption is ideal. A caption with high semantic alignment might still have non-essential elements, while focusing solely on essential elements might not adequately represent the image. To address this issue, we propose the \ourmethod{} framework, which uses scoring methods to evaluate both the degree of semantic alignment between the image and caption, and how essential the terms in the captions are.

To determine how well captions match images, we used various pre-trained CLIP \cite{radford2021learning} models and BLIP-2\cite{li2023blip} model to calculate the cosine similarity between image and text features. We then created a robust \ensclip{} by combining the results. To measure how essential the terms in the captions are, we used a \consensus{} derived from comparing candidates within the pool of captions. Finally, we combined the \ensclip{} and \consensus{} to calculate the final score. If the difference between the top two captions for an image is negligible, both captions are considered equally good at describing the image. In this case, we choose the caption with fewer words as the final caption.

The proposed \ourmethod{} framework is a method for caption evaluation that is easy to understand and doesn't require any additional fine-tuning. It can take into account both the alignment of images and text, as well as the essentialness of captions in a zero-shot setting. The overall framework can be seen in \cref{fig:pipeline}. By using this approach, we were able to achieve impressive results in the NICE 2024 Challenge. We came in third place based on the CIDEr metric, second place in both SPICE and METEOR metrics, and first place in the ROUGE-L and BLEU metrics. These achievements show that our method is not limited to excelling in a single metric, but is versatile and can be applied to various evaluation criteria.

\section{Proposed Method}
\label{sec:proposed_method}
\begin{figure*}
    \centering
    \includegraphics{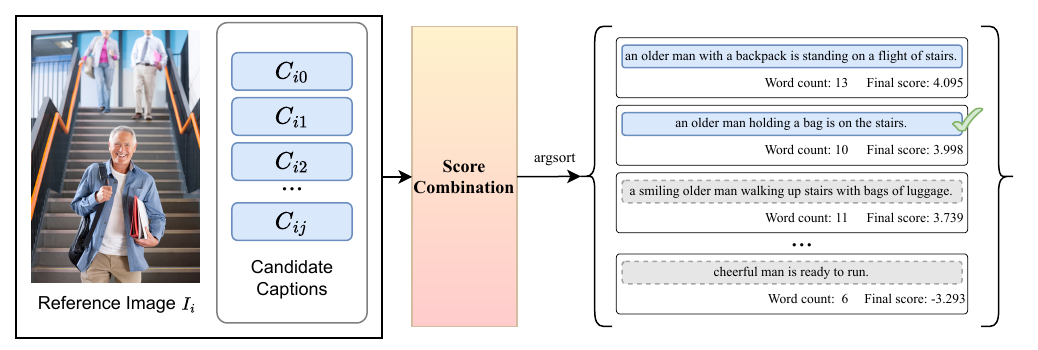}
    \caption{
The Short Cap Selection process: If the final scores of the top 2 captions differ by less than the threshold $\mathbf{\theta}$, a caption with fewer words is chosen.}
    \label{fig:short_cap_selection}
\end{figure*}

Building on the concept of a ‘well-explained image caption' defined in \cref{sec:intro}, we introduce \ourmethod{}, a framework designed to select the ideal caption by considering both the semantic alignment between images and captions and the essentialness of captions. \ourmethod{} comprises two main scoring algorithms: 1) the \ensclip{} and 2) the \consensus{}. In \cref{subsec:ens_clip} discusses the integrated CLIP score derived from the cosine similarity between image and caption embeddings, utilizing a variety of CLIP models and the ITC Loss calculation from BLIP-2 to assess the alignment between images and captions. In \cref{subsec:consensus} covers the method of measuring essentialness through mutual comparison between candidate captions, termed the \consensus{}. In \cref{subsec:score_comb_set}, we detail the process of integrating these two scores, and \cref{subsec:Shortcapselect} explains how the combined score is used to select the final caption.


\subsection{\ensclip{}}
\label{subsec:ens_clip}
The CLIP score\cite{hessel2022clipscore} is a metric that measures the semantic alignment between images and captions by comparing the cosine similarity between the image embedding $\mathbf{E}_{I}$ and caption embedding $\mathbf{E}_{C}$ through a pre-trained CLIP model. However, as the training data for the pre-trained CLIP model differs from the zero-shot caption re-ranking dataset provided, the accuracy of the single CLIP score may not be reliable. To address this issue, an ensemble of CLIP scores from various models that have proven to perform well in zero-shot tasks can provide a more robust semantic alignment than a single CLIP score.

\begin{equation}
S_{\text{CLIP}}(I, C) = \cos(E_I, E_C)
\end{equation}

\begin{equation}
S' = \frac{S - \text{mean}(S)}{\text{std}(S)} .
\end{equation}

\begin{equation}
\begin{aligned}
S_{\text{ensemble}} &= \sum_{i \in I} {S_{\text{CLIP}}^{i}}', \quad \forall i \in I, \\
\text{where } I &= \{ \text{“EVA-CLIP”, “MetaCLIP”, “MobileCLIP”,} \\
&\quad \text{“OpenCLIP”, “BLIP-2”} \}.
\end{aligned}
\end{equation}
The conventional CLIP score determines semantic alignment using the cosine similarity between $\mathbf{E}_{I}$ and $\mathbf{E}_{C}$, substituting any negative values with zero. However, the \ourmethod{} framework allows negative values to remain to achieve a more refined score distribution for image-caption pairs with low relevance. To calculate the \ensclip{}, the cosine similarity values between $\mathbf{E}_{I}$ and $\mathbf{E}_{C}$ were calculated using models such as EVA-CLIP-18B\footnote{\url{https://github.com/baaivision/EVA/tree/master/EVA-CLIP-18B}}, MetaCLIP\footnote{\url{https://github.com/facebookresearch/MetaCLIP}}, MobileCLIP\footnote{\url{https://github.com/apple/ml-mobileclip}}, OpenCLIP\footnote{\url{https://github.com/mlfoundations/open_clip}}, and BLIP-2\footnote{\url{https://github.com/salesforce/LAVIS/tree/main/projects/blip2}}.


\subsection{Consensus score}
\label{subsec:consensus}
We refer to the extent to which a caption is made up of essential expressions as its "Essentialness". When various models produce different captions, the expressions that appear most often are considered essential to describe the image. To measure this essentialness, we use a scoring method called the \consensus{}. 

The \consensus{} is a metric derived from the CIDEr score, that calculates the TF-IDF weights for N-Grams across candidate and reference captions. It then calculates the cosine similarity between the TF-IDF weight vectors of the candidate caption and each reference caption. To assess the essentialness of expressions within a caption, we calculate the \consensus{} for each caption, using all remaining candidate captions as reference captions, except the caption under evaluation.

However, the effectiveness of the \consensus{} is significantly influenced by the quality of the caption pool used as references. In other words, if the reference caption set consists only of high-quality captions, the \consensus{} is more likely to reliably reflect the degree of essentialness. To enhance the effectiveness of the \consensus{}, we use two filters to make a high-quality caption pool.

\subsection{Caption Filtering}
\label{subsec:caption_filter}
The consensus scoring system gives higher scores to captions that use essential words frequently used in multiple captions. However, if the pool of candidate captions has many irrelevant or non-conforming captions, this scoring method may not work. To solve this issue, we filter the candidate caption pool with two types of filtering.
\subsubsection{Bad Format Filter}
\label{subsec:badformat_filter}
Based on the insights from the Flickr30k \cite{young2014image} and COCO \cite{lin2014microsoft} datasets, we have identified the typical structure of captions. Generally, a caption is a phrase or clause of a single sentence and includes a sufficient amount of information in an image. To ensure high-quality and relevant captions, we filtered out captions that contained more than two periods or more than three commas, or those that had fewer than five words. This filtering was done systematically using a rule-based algorithm. This process helps to ensure that the captions being evaluated adhere to conventional standards.
\subsubsection{ITM Filter} 
\label{subsec:itm_filter}
In order to filter out captions that are irrelevant to the content of the image from group of candidates, we implement another filter which is called ITM filter. Removing captions that are not related to the image is important because they can hinder the consensus scoring. The consensus scoring is based on the agreement among captions and can be affected by the inclusion of expressions that are not related to the image content.

To filter out irrelevant captions, the Image-Text Matching (ITM) Loss from BLIP-2 is used. The ITM loss is designed to classify an image and text pair as either positive or negative, making it very efficient for filtering out captions that are not related to the image.

The ITM loss is calculated for each caption associated with an image, and the top 50\% of captions with the highest ITM values are selected. These captions are then used in the caption pool for consensus scoring, ensuring that the captions considered are more likely to be relevant and aligned with the image content.

 \begin{figure}
    \centering
    \includegraphics[width=0.9\linewidth]{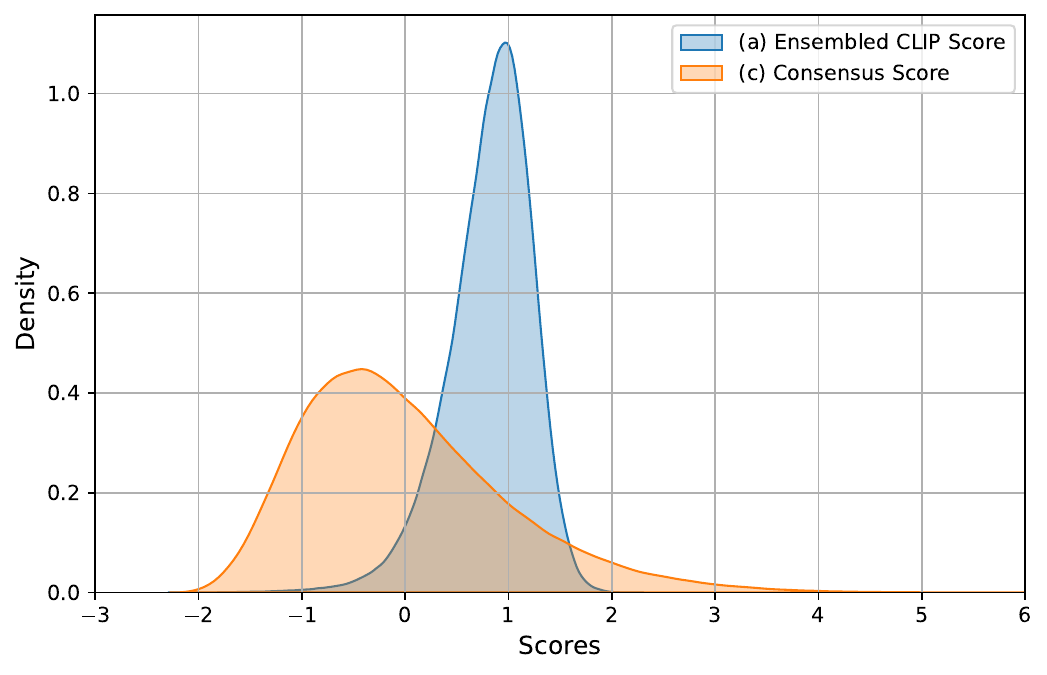}
    \caption{Comparison of the \ensclip{} and \consensus{} distributions}
    \vspace{-5pt}
    \label{fig:clip_con_distribution}
\end{figure}

\subsection{Score Combination}

In \cref{subsec:ens_clip} and \cref{subsec:consensus}, we defined the \ensclip{} and the \consensus{}. After normalizing these scores individually, we combine them using a weighted sum to form the final score. This approach allows us to adjust the influence of each score differently, ensuring that both the semantic alignment between the image and captions and the essentialness of the captions are appropriately considered in determining the most suitable caption. This method of integration provides a flexible framework that can be tailored to prioritize different aspects of caption quality depending on the specific requirements of the task at hand.
\begin{equation}
\begin{split}
S_{comb} = \lambda_1 S'_{ensemble} + \lambda_2 S'_{consensus} .
\end{split}
\end{equation}

\subsection{Short Caption Selection}
\label{subsec:Shortcapselect}
By combining the earlier \ensclip{} and \consensus{}, we obtained a final score that reflects both the semantic alignment between the caption and image, and the essentialness of the caption. If there is a clear distinction in the final score, the caption with the highest score is selected as the optimal caption. In cases where the difference in the final score is not pronounced, meaning the difference between the scores of the top-2 captions is less than a threshold $\mathbf{\theta}$, we chose the caption with fewer words as the final caption from the perspective of essentialness as shown in \cref{fig:short_cap_selection}.

\section{Experiments}
\label{sec:Experiments}

\subsection{Score Combination Setting}
\label{subsec:score_comb_set}

\begin{figure}
    \centering
    \includegraphics[width=0.8\linewidth]{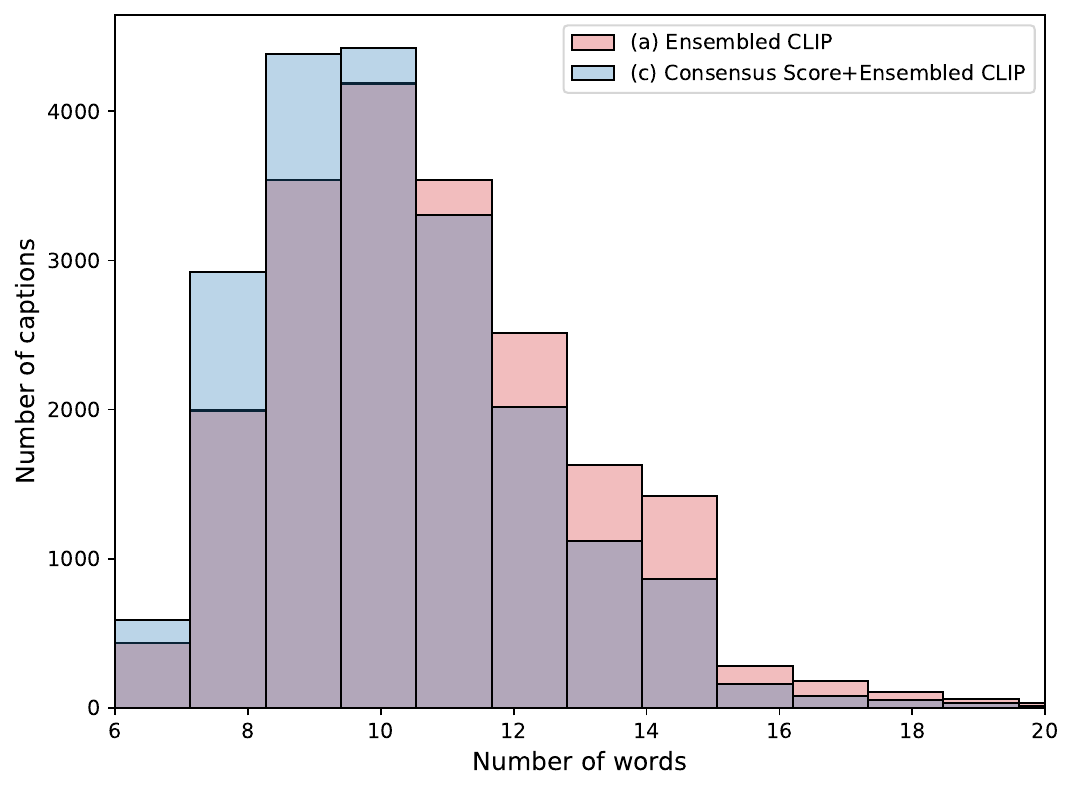}
    \caption{Comparison of the number of words in captions selected by the \ensclip{} versus the \consensus{}.}
        \vspace{-5pt}
    \label{fig:Number_of_word}
\end{figure}

\begin{table*}[t!]
    \centering
\begin{tabular}{@{}lcccccc@{}}
  \toprule
  Method & CIDEr & SPICE & METEOR & ROUGE-L & BLEU$_{\text{AVG}}$ \\
  \midrule
  \midrule 
  (a) Ensembled CLIP Score & 176.83 & 30.06 & 34.84 & 63.88 & 57.88 \\
  (b) Consensus Score & 202.89 & 31.21 & 35.79 & 68.20 & 67.07 \\
  (c) Ensembled CLIP Score + Consensus Score & \underline{212.44} & \underline{32.41} & \underline{36.98} & \underline{69.41} & \textbf{67.90} \\
  (d) $3.52$*Ensembled CLIP Score + Consensus Score & \textbf{218.47} & \textbf{33.46} & \textbf{37.98} & \textbf{70.13} & \underline{67.58} \\
  \bottomrule
\end{tabular}
    \caption{\textbf{Ablations of Score Combination}: The caption with the highest score is ultimately selected for submission. This involves \textbf{(a)} combining the CLIP score from models like EVA-CLIP-18b, MetaCLIP, MobileCLIP, and OpenCLIP with the BLIP-2 ITC score, \textbf{(b)} using consensus-based scoring alongside Caption Filtering, \textbf{(c)} combining the Ensembled CLIP score with the Consensus score at a 1:1 ratio, and \textbf{(d)} combining the Ensembled CLIP score with the Consensus score at a ratio of 3.52:1.}
    \label{tab:table1}
    \vspace{5pt}
\end{table*}

\begin{table*}[t!]
    \centering
    \resizebox{0.8\textwidth}{!}{ 
    \begin{tabular}{@{}lcccccccc@{}}
      \toprule
      Method & CIDEr & SPICE & METEOR & ROUGE-L & BLEU$_{\text{AVG}}$\\
      \midrule
      (b) Consensus Score & \textbf{202.89} & \textbf{31.21} & \textbf{35.79} & \textbf{68.20} & \textbf{67.07} \\
      (e) Consensus Score (w/o Caption Filtering) & 192.98 & 30.10 & 34.41 & 65.89 & 64.75 \\
      \bottomrule
    \end{tabular}
    }
    \caption{\textbf{Ablations of Caption Filtering.} \textbf{(e)} has the same settings as \textbf{(b)} with the exception of using a caption filter.}
    \label{tab:table2}
\end{table*}

\begin{table*}[t!]
    \centering
    \resizebox{0.8\textwidth}{!}{ 
    \begin{tabular}{@{}lcccccccc@{}}
      \toprule
      Method & CIDEr & SPICE & METEOR & ROUGE-L & BLEU$_{\text{AVG}}$\\
      \midrule
      (d) 3.52*Ensembled CLIP Score + Consensus Score & 218.47 & \textbf{33.46} & \textbf{37.98} & 70.13 & 67.58 \\
      (f) \hspace{8.7em} + Short Capation Selection & \textbf{220.53} & 33.01 & 37.68 & \textbf{70.31} & \textbf{69.20} \\
      \bottomrule
    \end{tabular}
    }
    \caption{\textbf{Ablations of Short Caption Selection.} \textbf{(f)} has the same settings as \textbf{(d)} with the addition of Short Caption Selection. We set the threshold $\mathbf{\theta}$ to 0.39.} 
    \label{tab:table3}
\end{table*}

When setting the weights for score combination, we observed significant differences in outcomes depending on how the \consensus{} and the \ensclip{} score were utilized. When comparing selected captions by using only the \consensus{} to those by using \consensus{}and \ensclip{} equally, 
We find a difference in 5,396 out of 20,000 captions. Conversely, the discrepancy reached 18,217 captions when the \ensclip{} was used alone versus when it was combined with the \consensus{}. To analyze these differences accurately, we visualize the distribution of both scores after normalization and discovered that the maximum value of the \consensus{} was approximately three times larger than that of the clip score as shown in \cref{fig:clip_con_distribution}. This discrepancy suggested that, in situations where the caption with the highest combined score was selected, the overwhelming influence of the \consensus{} could skew the results.

To ensure a balanced reflection of both scores, we decided to set $\mathbf{\lambda}{_{1}}$ (the weight for the \ensclip{}) larger than $\mathbf{\lambda}{_{2}}$ (the weight for the \consensus{}). After a few experiments, we confirm that a ratio of 3.52:1 is the most effective. This decision is supported by experimental evidence presented in \cref{tab:table1}, where the CIDEr score for results combined equally is 212.44, compared to 218.46 for combinations using the 3.52:1 ratio. This result confirm that placing greater weight on the $\mathbf{\lambda}{_{2}}$ leads to improved outcomes. This weighting strategy aims to balance the influence of both the \consensus{} and the \ensclip{}, ensuring that both semantic alignment and essentialness are appropriately considered in the final caption selection.

\subsection{Consensus Scoring's Effectiveness in Identifying Essentialness}
\label{subsec:consensus_essential}

We conduct an evaluation of the effectiveness of using the ITM Filter and Bad Format Filter, by comparing \consensus{}{} for captions with and without filtering. Based on the results presented in \cref{tab:table2}, we find that the filtered case has a \consensus{} of 202.89, while the unfiltered case has a score of 192.98. This indicates that filtering the caption pool improves the quality of captions selected, as measured by the CIDEr metric.

Furthermore, we created a visualization in \cref{fig:Number_of_word} to show the length of captions selected from each pool, measured by the number of words per caption. The visualization shows that captions chosen from the filtered pool are significantly shorter on average within their respective pools. These findings collectively suggest that filtering the caption pool enhances the ability of consensus scoring to assess essentialness. By improving the overall quality of captions selected, this strategy maximizes the functionality of consensus scoring.

\subsection{Effects of Caption Filtering}
\label{subsec:CaptionFilterablation}

\begin{figure}
    \centering
    \includegraphics[width=0.8\linewidth]{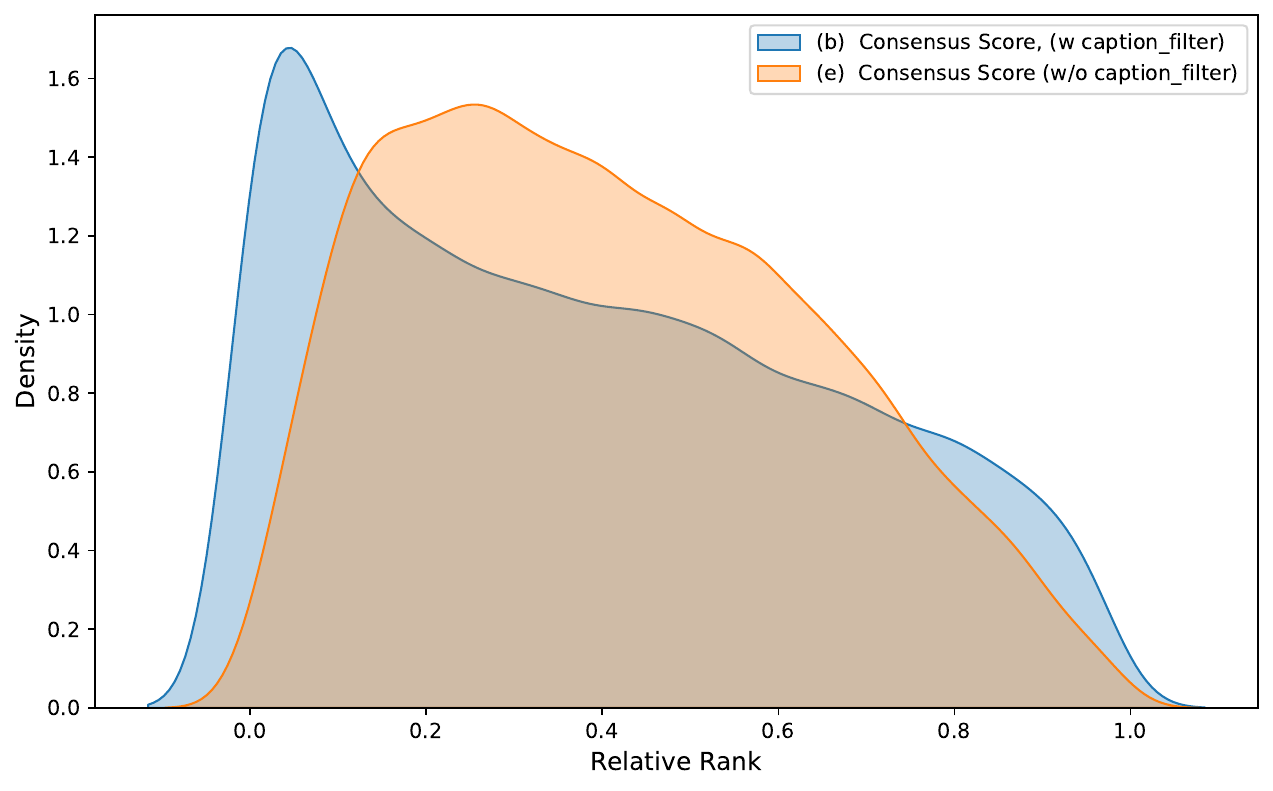}
    \vspace{-7pt}
    \caption{The relative rank of the selected caption within the candidate caption pool in terms of the number of words.}
    \label{fig:caption_filter_ ablation}
\end{figure}

To assess the effectiveness of the ITM Filter and Bad Format Filter, we compare the evaluation results of the filtered cases with those of the unfiltered cases. The results, \cref{tab:table2}, shows that the CIDEr score increases from 192.98 to 202.89 and there are also improvements in every other metrics. It demonstrate an enhancement in consensus scoring by refining the pool of captions. Furthermore, we visualized the relative rank of the selected caption within the candidate caption pool in terms of the number of words. \cref{fig:caption_filter_ ablation} reveals that, after filtering the pool, the chosen captions are significantly shorter than those in their respective pools. These findings collectively suggest that filtering the caption pool enhances the ability of consensus scoring to discern essentialness, maximizing its effectiveness in evaluating captions. 

\subsection{Effects of the Short Caption Selection}
\label{subsec:ShortCapablation}

When comparing the results of applying Short Caption Selection to those without it, as shown in \cref{tab:table3}, it's evident that using Short Caption Selection improves performance: the CIDEr score increased from 218.47 to 220.53 upon application. Also there are improvement in some other metrics (ROUGE-L, BLEU) when Short Caption Selection was applied, compared to when it was not. This shows that choosing shorter captions can make the caption selection process more effective.

 \begin{figure}
    \centering
    \includegraphics[width=1\linewidth]{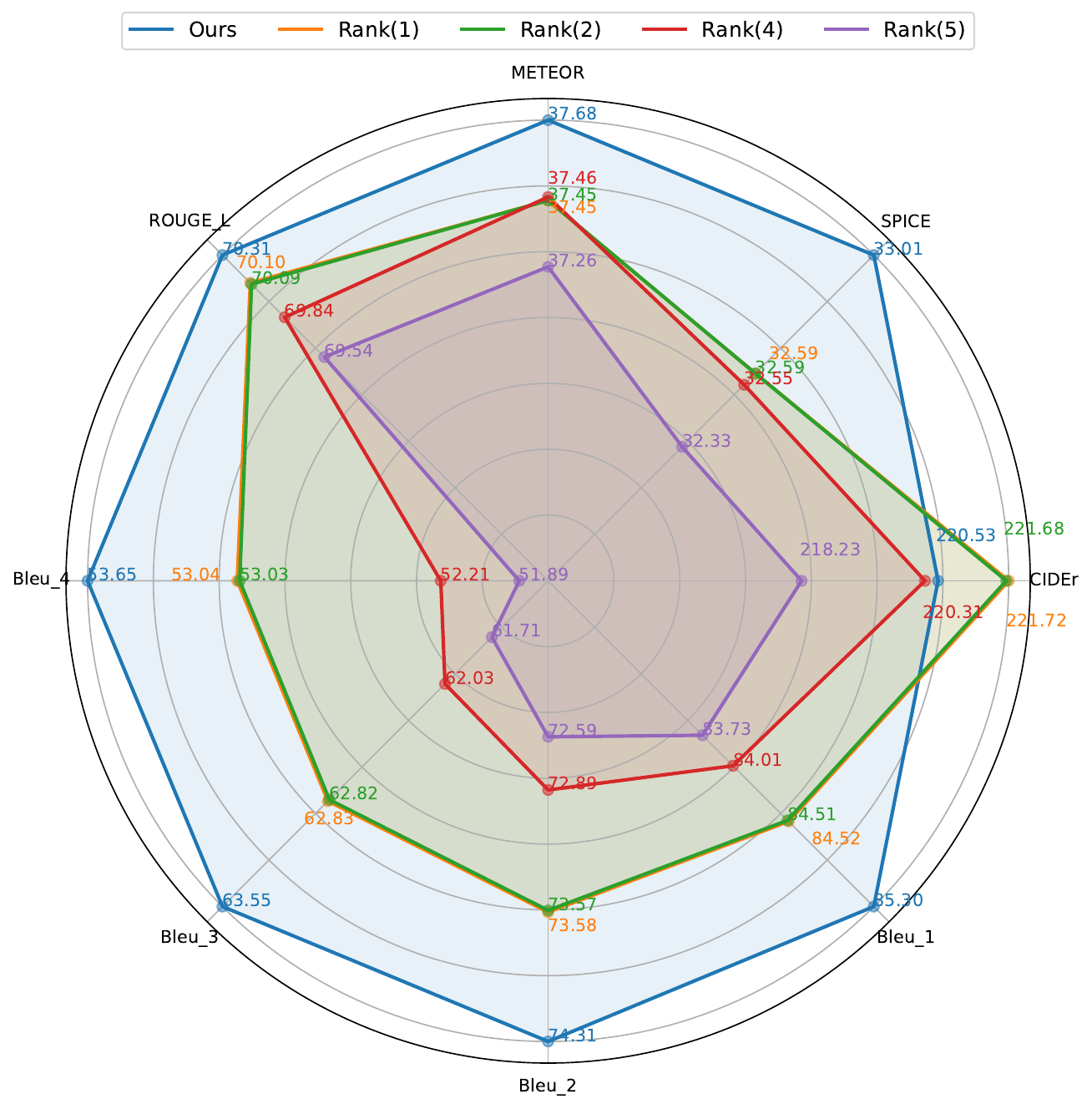}
    \vspace{-10pt}
    \caption{Comparative advantage of our methodology. Our approach demonstrates a marked improvement over other top-ranked methodologies.}
    \label{fig:radarchart}
\end{figure}
\section{Conclusion}
\label{sec:Conclusion}

We propose \ourmethod{}, a zero-shot caption re-ranking framework that incorporates both image-caption semantic alignment and caption essentialness. Our method selects the most ideal caption for an image from several candidates, without model training. Through \cref{fig:radarchart}, we have verified that our methodology serves as a general caption re-ranking framework that performs well across all metrics, demonstrating its effectiveness and versatility in identifying ideal captions.
{
    \small
    \bibliographystyle{ieeenat_fullname}
    \bibliography{main}
}

\end{document}